\documentclass{article}

\usepackage[preprint,nonatbib]{neurips_2021}
\usepackage[numbers]{natbib}
\usepackage[utf8]{inputenc} % allow utf-8 input
\usepackage[T1]{fontenc}    % use 8-bit T1 fonts
\usepackage{hyperref}       % hyperlinks
\usepackage{url}            % simple URL typesetting
\usepackage{booktabs}       % professional-quality tables
\usepackage{amsfonts}       % blackboard math symbols
\usepackage{nicefrac}       % compact symbols for 1/2, etc.
\usepackage{microtype}      % microtypography

\usepackage{tablefootnote}
\usepackage{graphicx}
\usepackage{amsmath}
\usepackage{amssymb}
\usepackage{subfigure}
\usepackage{amsfonts}
\usepackage{amssymb}
\usepackage{bbm}
\usepackage{amsthm}

\usepackage{extarrows}
\usepackage{bm}
\usepackage{multirow}
\usepackage[dvipsnames]{xcolor}
\usepackage{multicol}

\usepackage{authblk}

\title{Message Passing in Graph Convolution Networks via Adaptive Filter Banks}

\author[$\ddagger$*]{Xing Gao}

\author[$\diamond$]{Wenrui Dai}
\author[ $\ddagger$]{Chenglin Li}
\author[$\diamond$]{Junni Zou}
\author[$\ddagger$]{Hongkai Xiong}
\author[$\dagger$]{Pascal Frossard}
\affil[$\ddagger$] {Department of Electronic Engineering, \ Shanghai Jiao Tong University}
\affil[$\diamond$] {Department of Computer Science, \ Shanghai Jiao Tong University}
\affil[$\dagger$] {Signal Processing Laboratory (LTS4), \ EPFL}
\affil[*]{william-g@sjtu.edu.cn}
\begin{document}

\maketitle

\begin{abstract}
Graph convolution networks, like message passing graph convolution networks (MPGCNs),  have been a powerful tool in representation learning of networked data. However, when data is heterogeneous, most architectures are limited as they employ a single strategy to handle multi-channel graph signals and they typically focus on low-frequency information.   In this paper, we present a novel graph convolution operator, termed BankGCN, which keeps benefits of message passing models, but extends their capabilities beyond `low-pass' features. It decomposes multi-channel signals on graphs into subspaces and handles particular information in each subspace with an adapted filter. The filters of all subspaces have different frequency responses and together form a filter bank. Furthermore,  each filter in the spectral domain corresponds to a message passing scheme, and diverse schemes are implemented via the filter bank. Importantly, the filter bank and the signal decomposition are jointly learned to adapt to the spectral characteristics of data and to target applications. Furthermore, this is implemented almost without extra parameters in comparison with most existing MPGCNs. Experimental results show that the proposed convolution operator permits to achieve excellent performance in graph classification on a collection of benchmark graph datasets. 
\end{abstract}

\section{Introduction}
In many scientific and application domains, data is supported by a graph or network structure. To deal with such data, a collection of graph convolution networks (GCNs) have been proposed by generalizing architectures from the Euclidean domain to the graph one. Among them, MPGCNs \citep{bruna2013spectral,gilmer2017neural,velikovi2018graph,hamilton2017inductive,xu2018how,klicpera2019diffusion} built on diverse MP schemes are more prevalent with their  flexible and intuitive formulations of graph convolution, and have presented state-of-the-art performance in a series of tasks,  like node classification and graph classification. 

The message passing schemes in MPGCNs, however,  mostly focus on the low frequency characteristics of the data. For example, GCN \citep{kipf2016semi} performs as Laplacian smoothing \citep{li2018deeper}, and the mean aggregator adopted in GraphSage \cite{hamilton2017inductive} has naturally low-pass properties. However,  in addition to low-frequency components, middle- and high-frequency components that correspond to the large variation of signals may also contain rich information in graph data. Besides, given that the importance of different frequency components may vary with the tasks,  it is beneficial to be able to exploit more than the low-frequency information in an adaptive way. Furthermore, MPGCNs usually adopt a single message passing and aggregation scheme. This limits the capability of MPGCNs to handle graphs whose features or signals, like in most applications,  are multi-channel and encode heterogeneous information with different frequency characteristics. For instance, node features correspond to user profiles in social networks may include gender, age, hobby, etc., and the gender and age features will definitely present different variations across the graph.  

\begin{figure}[tp]
\vskip 0.2in
\begin{center}
\centerline{\includegraphics[width=0.99\columnwidth]{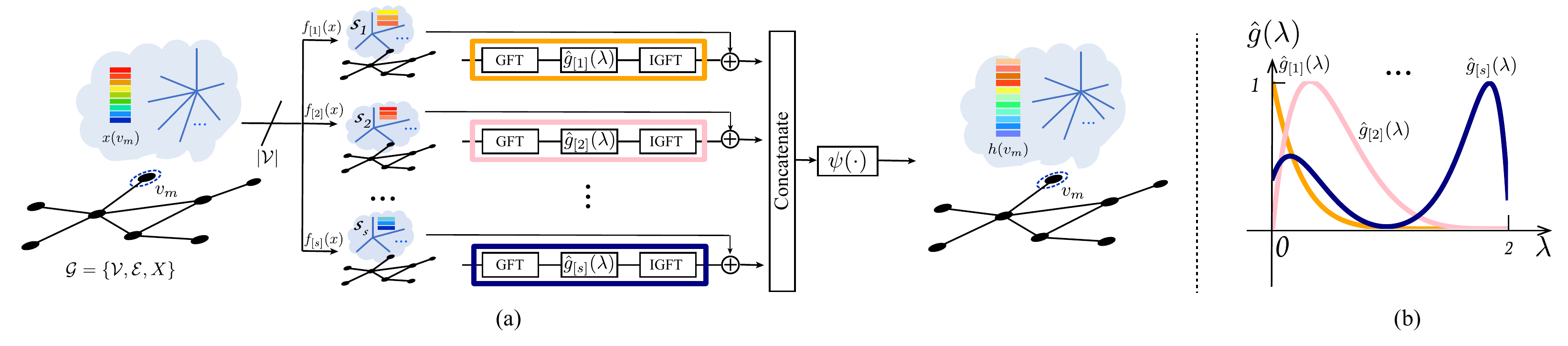}}
\caption{Illustration of the framework. The proposed convolution operator and its equivalent filter banks are respectively presented in (a) and (b). Graph signal $\bm{x}$ is mapped to a collection of subspaces and processed by an adaptive filter bank to extract information corresponding to distinct frequency properties. As illustrated in (b), an adaptive filter bank permits the proposed convolution operator to adaptively extract information from diverse combinations of different frequency components to produce graph representations.}
\label{fig.frame}
\end{center}
\vskip -0.2in
\end{figure}

In order to address these issues, we propose a novel message passing graph convolution operator termed BankGCN that utilizes an adaptive and learnable filter bank to process heterogeneous graph data in the  spectral  (frequency) domain,  as presented in Fig.~\ref{fig.frame}. Firstly, we decompose multi-channel graph signals into a collection of subspaces through projection, in order to separate input data according to its spectral characteristic. In each subspace, a learnable filter is employed to capture the frequency properties of the particular signals. Notably, the filter is based on the universal design \citep{tremblay2018design} that is defined over a continuous range in the spectral domain  instead of a specific discrete spectrum, and thereby are adaptable to graphs with arbitrary topologies \citep{levie2019transferability}. Besides, it is designed as a finite impulse response (FIR) filter and corresponds to a local message passing scheme in the spatial domain that jointly considers multi-hop topological information and signal information. Filters of all the subspaces together form a filter bank, and they are simultaneously learned from data together with subspace projections. Furthermore, a diversity condition is proposed to regularize the filters in the filter bank to have diverse frequency responses, in order to capture and handle various patterns in the graph signal. In this way, BankGCN is more powerful than most MPGCNs, such as GCN \citep{kipf2016semi}, GraphSage \citep{hamilton2017inductive}, and GIN \citep{xu2018how}, in that it is able to exploit more than `low-pass' features in the data and adapts to its heterogeneous properties.  It further adopts a group of learnable message passing strategies, and exploits multi-hop rather than one-hop information per layer. In contrast with spectral methods, like ChebNets \citep{defferrard2016convolutional} and CayleyNets \citep{levie2018cayleynets}, it largely reduces the number of free parameters and leads to better generalization in the experimental validation. The proposed convolution operator is stackable,  as most MPGCNs, and can be optimized together with other modules in the GCNs like graph  pooling.  In this paper, we evaluate it in the task of graph classification on a collection of benchmark datasets, where it achieves superior performance. Remarkably, it is promising to extend BankGCN to tasks like link prediction and applications into non-Euclidean data like 3-D point clouds.

The remainder of the paper is organized as follows. Section~\ref{sec.related} briefly overviews the related work on graph convolution and spectral filtering. In Section~\ref{sec.pb}, some preliminaries are introduced and a collection of graph convolution operators are compared in terms of spectral filtering. Section~\ref{sec.method} elaborates the proposed BankGCN algorithm. Evaluations on graph classification tasks are presented in Section~\ref{sec.eval}. Finally, Section~\ref{sec.con}  concludes this paper.

\section{Related Work}\label{sec.related}
We briefly overview below several spatial GCNs in terms of their respective message schemes. Then we introduce spectral filtering as well as the design of filters and filter banks in graph signal processing (GSP), and compare several spectral GCNs.

\textbf{Message Passing Graph Convolution Networks.} Several MPGCNs \citep{bruna2013spectral,gilmer2017neural,velikovi2018graph,hamilton2017inductive,xu2018how,klicpera2019diffusion} have been proposed to generalize convolution operation to graph data with a variety of message propagation and aggregation schemes in the spatial domain. For instance,  message are aggregated with the node-wise mean or max in a localized neighborhood in GraphSage \citep{hamilton2017inductive},  or based on attention scores in GAT  \citep{velikovi2018graph}.  A more expressive scheme, GIN, is further proposed with summing features in a neighborhood followed by a multi-layer perceptrons (MLP) to approximate any injective function on multiset in \citep{xu2018how}. However, these methods are constrained to a single message passing (MP) strategy, mostly capturing `low-pass' characteristics, and the MP range is usually of one-hop neighborhood per layer. Focusing on these issues,  \citeauthor{klicpera2019diffusion} expand the MP range to multi-hop neighborhoods by adopting graph diffusion convolution (GDC) \citep{klicpera2019diffusion}; diffusion wavelets are introduced in Scattering GCN \citep{MinWW20} from geometric scattering transform \citep{gao2019geometric} to complement the `low-pass' features of GCN \citep{kipf2016semi} with `band-pass' features; and multiple aggregations, like mean, sum, and standard variation, are used to aggregate neighborhood information in \citep{corso2020principal}. In contrary to these methods,  we employ an adaptive and learnable filter bank rather than predefined wavelets and impose a proper decoupling between sets of learnable filters, which correspond to different MP schemes within multi-hop neighborhoods.

\textbf{Filtering on Graphs.} Frequency filtering, or spectral filtering, is generalized to graph data via spectral graph theory \citep{shuman2013emerging}. Correspondingly, several spectral GCNs are derived with graph filters, or equivalently graph convolutions,  designed in graph spectral domain directly, as pioneered by \citet{bruna2013spectral}.  To achieve constant learning complexity and avoid expensive eigendecomposition of the graph Laplacian,  similarly to dictionary learning methods \citep{thanou2014learning} in GSP, the filters are adopted as specific functions of eigenvalues of the graph Laplacian,  such as Chebyshev polynomials in ChebNets \citep{defferrard2016convolutional} and TIGraNet \citep{khasanova2017graph}, Cayley polynomials in CayleyNets \citep{levie2018cayleynets}, and auto-regressive moving average (ARMA) filters in \citep{bianchi2019graph}. In contrast with these spectral convolution methods that focus on the implementation of filters with desirable properties, like localization and narrowband specialization,  our paper focuses on the design of the filter bank, whose filters can be built on polynomial kernels, in order to reduce the number of filters and thereby the number of parameters, and improve the generalization.  

Finally, there are several papers in graph signal processing literature about the design of graph filter banks for graph signal decomposition \citep{narang2012perfect,tanaka2014m} and multiscale analysis \citep{hammond2011wavelets,narang2013compact}. However, they usually work under a rigid constraint, perfect reconstruction, and thereby their produced (sparse) representations may not be flexible enough to adapt to diverse tasks in machine learning. In contrast, our method relaxes the perfect reconstruction condition and rather regularizes filters in a filter bank to be different in the spectral domain.  It further learns the filters and subspace projections jointly through end-to-end optimization in GCNs to flexibly handle multi-channel signals.

\section{Preliminaries and Problems} \label{sec.pb} 
In this paper, we consider data that is represented on undirected graphs. An input graph is described as $\mathcal{G}=(\mathcal{V},\mathcal{E})$, with $\mathcal{V}$ and $\mathcal{E}$ respectively denoting the set of vertices and the set of edges.  We generally use capital letters for matrices and bold lowercase letters for vectors. The graph topology is characterized by the adjacency matrix $A$ with a non-zero value $(A)_{ij}$ indicating the weight of the edge connecting vertices $v_i$ and $v_j$.  If available \footnote{If node attributes are not available, we adopt the structural information like the one-hot encoding representation of node degree  and the local clustering coefficient as node signals.}, the $d$-channel signals or attributes of nodes are represented as $\bm{x}(v_{m}) \in \mathbb{R}^{d}$  for $\forall v_m \in \mathcal{V}$, and $X=[\bm{x}(v_{1}),\bm{x}(v_{2}), \cdots, \bm{x}(v_{n})]^T$ is for the whole graph $\mathcal{G}$ with $n=| \mathcal{V}|$ nodes.  Specially, for vector features, $x_i(v_{m}) $ indicates the $i$-th channel of the signal on vertex $v_m$, and $\bm{x_i}=[x_i(v_{1}),x_i(v_{2}), \cdots, x_i(v_{n})]^T$ represents this signal on the whole graph. For the hierarchical structure of GCNs, we employ a subscript $l$ to indicate variables or parameters belonging to the $l$-th layer. In most cases, we ignore $l$ for clarity without causing confusion. 

The Graph Fourier Transform (GFT) is defined based on the graph Laplacian matrix $L$ \citep{shuman2013emerging}. We adopt the symmetric normalized graph Laplacian operator, \emph{i.e.}, $L=I- D^{-\frac{1}{2}}AD^{-\frac{1}{2}}$ with the diagonal degree matrix $D$ defined as $(D)_{ii}=\sum_{j}(A)_{ij}$. The eigendecomposion of the Laplacian matrix is denoted by $L=U \Lambda U^{*}$, where $U=\lbrack \bm{u_1},\bm{u_2}, \dots, \bm{u_{n}} \rbrack$  is composed of the eigenvectors $\bm{u_k}$, $k=1,\cdots,n$ and $ \Lambda$ is a diagonal matrix with  eigenvalues $\lambda_1,\lambda_2,\dots,\lambda_{n}$, with $0=\lambda_1 \leq  \lambda_2  \leq \dots \leq   \lambda_{n}\leq 2$. The eigenvalues $\{ \lambda_k\}$ construct the spectrum of the graph $\mathcal{G}$ \footnote{In analogy to classical Fourier analysis, eigenvalues provide a corresponding notion of frequency and lead to frequency filtering.}, and the GFT of a signal $\bm{x}$ is calculated as the inner-product between each of its component $\bm{x_i}$ and the eigenvectors $\{\bm{u_k}\}$ \citep{shuman2013emerging}:

\begin{equation}\label{e.gft}
\hat{x}_i(\lambda_k)=\sum_{m=1}^{n}x_i(v_m)u_k^*(v_m),
\end{equation}
where $*$ indicates conjugate transpose and $u_k^*(v_m)$ denotes the $m$-th element in $\bm{u}^*_k$. The Inverse Graph Fourier Transform (IGFT) is defined as
\begin{equation}\label{e.igft}
	x_i(v_m)=\sum_{k=1}^{n} \hat{x}_i(\lambda_k)u_k(v_m).
\end{equation}
According to the convolution theorem, the convolution between the $i$-th channel of the signal ${x}_i(v_m)$ and the corresponding filter $g(v_m)$ (with frequency response $\hat{g}(\lambda_k)$ ) is defined as:
\begin{equation}\label{e.conv}
	(x_i \ast g)(v_m)=\sum_{k=1}^{n}\hat{x}_i(\lambda_k)\hat{g}(\lambda_k)u_k(v_m)
\end{equation}
Let us define $\hat{g}(\Lambda)={\rm diag}([\hat{g}(\lambda_1), \hat{g}(\lambda_2), \dots, \hat{g}(\lambda_n)])$. We have for $\bm{x}_i(v_m)$ that
\begin{equation}
	(\bm{x_i} \ast g)=U\hat{g}(\Lambda)U^*\bm{x_i}.
\end{equation}

Finally, a graph filter bank is composed of a set of filters $\{\hat{g}_i(\Lambda)\}$ to decompose a graph signal into a series of signals with different frequency components \citep{narang2012perfect}.

We now show how filtering can be implemented by messaging passing, which is used in many state-of-the-art graph representation learning methods.  Filtering in the frequency domain and MP in the spatial domain is closely related  \citep{shuman2013emerging}. Taking the most popular GCN \citep{kipf2016semi} as an example, according to its initial formation, the $j$-channel of a filtered signal is
\begin{equation}\label{e.mpfrec}
	\bm{h_j}=\sum_{i=1}^{d_l}(I+D^{-\frac{1}{2}}AD^{-\frac{1}{2}})\Theta \bm{x_i}=\sum_{i=1}^{d_l}(2I-L)\Theta_{i,j}\bm{x_i}=\sum_{i=1}^{d_l}\Theta_{i,j}U\hat{g}(\Lambda)U^*\bm{x_i} 
\end{equation}
where $\Theta$ denotes learnable parameters, and $\hat{g}(\Lambda)=2-\Lambda$ is a low-pass filter that is consistent with the analysis in \citep{nt2019revisiting,MinWW20}. The renormalized version of GCN and GIN can be formulated similarly. Please refer to \citep{balcilar2021analyzing} for their specific equivalent supports of $\hat{g}(\Lambda)$. Besides, GraphSage uses the mean or max operator to aggregate information in neighborhoods. Although it is hard to explicitly formulate its corresponding frequency response, the element-wise mean operation is a low-pass operation by nature and the element-wise max operator performs as an envelope extraction that suppresses the high-frequency components.

Another line of works, spectral GCNs,  are directly derived from spectral filtering. For ChebNets and CayleyNets as representatives of spectral GCNs, the filters are  defined in the frequency domain as $\hat{g}(\lambda)= \sum_{k=0}^{K} \theta^{(k)}T_{k}(\lambda)$ using the Chebyshev polynomial or Cayley polynomial basis $\{T_{k}(\cdot)\}$ and correspondingly
\begin{equation}\label{e:conv.spec}
	\bm{h_j}=\sum_{i=1}^{d_l} U\hat{g}_{ij}(\Lambda)U^*\bm{x_i}=\sum_{i=1}^{d_l} U\sum_{k=0}^{K} \Theta^{(k)}_{i,j}T_{k}(\Lambda) U^*\bm{x_i},
\end{equation}
where $K$ is the order of the polynomial filters and $\{ \Theta^{(k)}\}$ are learnable parameters. 

Notably, the above MPGCNs employ just one single filter to handle all the channels of signals for each input graph, and then takes different (linear) combinations of the filtered signals to obtain the output signals of different channels. Thereby, most MPGCNs are restricted in the number and types of filters and have a limited capacity from a spectral perspective. On the contrary, ChebNets and CaylayNets adopt  $d_l \times d_{l+1}$ ($d_l$, $d_{l+1}$ denote the respective number of channels of input and output signals) different  $K$-order polynomial filters, as presented in Eq.~\eqref{e:conv.spec}. The number $K$ determines the order and thereby the capacity of filters. Spectral convolution operators employ such a large number of learnable filters that are powerful to handle various graph signals but at the expense of numerous parameters to learn. This will lead to overfitting.

Based on the analysis of the limitations of existing spectral and message passing graph convolutions, we focus now on the following problems for designing the message passing scheme and graph convolution operator.
\begin{itemize}
\item Q1: How to design a message passing scheme that can capture diverse features beyond mere low frequency components for different machine learning tasks like graph classification?
\item Q2: How to design a graph convolution operator that is capable to handle heterogeneous node features adaptively and differently without introducing excessive extra parameters?
\end{itemize}

\section{The BankGCN Algorithm}\label{sec.method}

In this section, we first outline the main elements of BankGCN. Subsequently,  the role of subspace projections, the design of filters in the filter bank, and the regularization for the filter bank to guarantee diversity are introduced in detail.
 
\subsection{Graph Convolution with Adaptive Filter Banks}
Multi-channel graph signals (features) are usually composed of multiple patterns, and different signal channels vary differently over nodes, leading to diverse frequency characteristics. To handle the multi-channel signals, we employ a filter bank composed of a set of filters with different frequency responses in the design of BankGCN. Furthermore, subspace projection is adopted to decompose the multi-channel information into several components with similar spectral properties.

For a multi-channel input node signal $\bm{x}(v_{m})$, we first adaptively decompose it into different subspaces and then employ different filters to deal with the signal components  in each subspace separately. The decomposition aims to map the components of signals with similar frequency characteristics into the same subspace in order to facilitate the subsequent adaptive filtering. This is implemented with learnable subspace projections.  Mathematically, the input signal $\bm{x}(v_{m})$ is projected into $s$ subspaces with a group of projection functions denoted by $\{f_{[p]}(\cdot)\}$,
\begin{equation}
\bm{r}_{[p]}(v_{m})=f_{[p]}(\bm{x}(v_{m})), \quad p=1, 2, \dots, s, \  \forall v_m \in \mathcal{V}.
\end{equation}  
Here, the subscript ``$[p]$'' indicates the terms belonging to the $p$-th subspace and $\bm{r}_{[p]}(v_{m})$ is the projected signal. The choices for projection functions will be introduced in Section~\ref{sec.subpro}. Subsequently,  an adaptive filter $g_{[p]}(\cdot)$ is employed to learn and represent the sepectral components independently in each subspace,
\begin{align}
\bm{h}_{[p]}(v_{m})&=(g_{[p]} \ast \bm{r}_{[p]}) (v_{m}),
\end{align}
where  we reuse `*' to denote the convolution between each channel of signal $\bm{r}_{[p]}$ and the filter $g_{[p]}$. 

The filtered signals $\bm{h}_{[p]}(v_{m})$, $p=1,2,\cdots,s$, of all the subspaces are concatenated to produce the output features. A Rectified Linear Unit (ReLU) is then used as a non-linear activation function,
\begin{gather}
\bm{h}(v_{l,m})={\rm Concat}(\bm{h}_{[1]}(v_{l,m}),\bm{h}_{[2]}(v_{l,m}), \dots, \bm{h}_{[s]}(v_{l,m})), \\
	\bm{x}(v_{l+1,m})=\text{ReLU}(\bm{h}(v_{l,m}),
\end{gather}
where the subscript $l$ indicates variables or parameters in the $l$-th layer to describe the forward propagation between different layers in a hierarchical architecture.

Specially, considering that some filters are only supported on medium-to-high frequency components that mean large signal variations in the graph domain,  we further introduce a shortcut  in each subspace corresponding to full-pass in the frequency domain in order to make the filtered signals stable. Correspondingly, the filtered signal $\bm{h}_{[p]}(v_{m})$ in each subspace becomes
\begin{equation}
	\bm{h}_{[p]}(v_{m})=(g_{[p]} \ast \bm{r}_{[p]}) (v_{m}) +\bm{r}_{[p]}(v_{m}).
\end{equation}
The shortcut can further facilitate back propagation of gradients, as for CNNs on grid-like data \citep{he2016deep}.   

The proposed graph convolution operator is stackable, and the subspace mapping function in the upper layers will further combine features from different subspaces in the preceding layers to enable information interaction between different channels. 

\subsection{Subspace Projection}\label{sec.subpro}
For the projection function $f_{[p]}(\cdot)$,  there is a variety of design choices. Here, we take the linear mapping as an example, since it is simple yet able to separate and recombine different channels of signals.   Specifically, the $d$-channel \footnote{ For the input signal with  $d=1$, we set $s$ as the feature channels in the neural network in the first layer so that the signal is scaled differently in each subspace. In the following hidden layers, it can be used as the other cases  to handle multi-channel features produced by previous layers  in GCNs. } signal $ \bm{x}(v_m)$ is projected into $s$ different subspaces with learnable matrices $\{W_{[p]}\}_{p=1}^{s}$.  For the sake of simplicity, all the subspaces have the same dimension in this paper. To increase flexibility, we further introduce a learnable bias $\bm{b}_{[p]}$ for each subspace. 
\begin{gather}\label{e.linmap}
	\bm{r}_{[p]}(v_m)=f_{[p]}(x(v_m))=W_{[p]}^T \bm{x}(v_m)+\bm{b}_{[p]}, \quad p=1, 2, \dots, s,
\end{gather}

The introduction of subspace projection brings in two advantages: (i) it simplifies the learning process. Since the multi-channel graph signals have been decomposed, the filter in each subspace just needs to learn to capture frequency characteristics of the corresponding signal components; 
(ii) through projecting signals into low-dimensional subspaces, it limits the dimension of output features that are produced by the adaptive filter bank, and thereby reduces free parameters and computations in the following layers.

\subsection{Design of Filters}\label{sec.filter}
In order to handle graphs with arbitrary topologies and diverse signals, we adopt universal and adaptive filters to construct the filter bank. 

For any function $t: \mathcal{D}=[0,2] \rightarrow \mathbb{R}$, we can obtain a corresponding filter whose frequency response is $\hat{g}_{[p]}(\lambda)=t(\cdot)$, and its spatial construction is computed through IGFT:
\begin{equation}
	g_{[p]}(v_m)=\sum_{i=1}^{n} \hat{g}_{[p]}(\lambda_i)u_i(v_m),
\end{equation} 
with $\bm{u_i}$, $i=1,\cdots,n$, the eigenvectors of the symmetric normalized graph Laplacian of a graph. Notably, we directly design the frequency response of the filter for the continuous range $\mathcal{D}=[0,2]$ in which the spectrum of an arbitrary graph locates as introduced in  Section~\ref{sec.pb}. In other words, given the discrete spectrum $\{\lambda_0, \lambda_1, \dots,   \lambda_n\}$ of an arbitrary graph,  we have $\lambda_i  \in  \mathcal{D}$  and obtain the corresponding filter value on these specific discrete values $\hat{g}_{[p]}(\lambda_i)=t(\lambda_i)$, for $i=1, 2, \dots, n$. Thereby,  the filter $g_{[p]}(\cdot)$  is adaptable to any graph even with different topologies, \emph{i.e.,} it is a universal form \citep{tremblay2018design,levie2019transferability}.

However, with such graph filters, the computation of graph convolution (Eq.~\eqref{e.conv}) needs computation-intense eigendecomposition of the graph Laplacian. To reduce the computational complexity, we constrain the filter to the $K$-order polynomial function space, similarly to parametric dictionary learning \citep{thanou2014learning} in GSP. It corresponds to an FIR filter. Mathematically, the frequency response of 
a filter can be represented as
\begin{equation}\label{e.filter}
	\hat{g}_{[p]}(\lambda)=\sum_{k=0}^{K} \alpha^{(k)}_{[p]}T_{k}(\lambda),
\end{equation}
where $\{T_{k}\}$ denotes a specific polynomial basis such as Chebyshev polynomials, and $\{\alpha^{(k)}_{[p]}\}$ indicates corresponding coefficients. With $\{\alpha^{(k)}_{[p]}\}$ learnable, we obtain an adaptive filter whose frequency response adapts to the data and to the target task. Correspondingly, for the signal projected to the $p$-th subspace  $\bm{r}_{[p]}(v_m)$ , the filtered signal is calculated as
\begin{equation}\label{e.adpconv}
 \bm{h}_{[p]}(v_m)=(\bm{r}_{[p]} \ast g_{[p]})(v_m)=\sum_{i=1}^{n}u_i(v_m)\hat{g_{[p]}}(\lambda_i) \bm{\hat{r}}_{[p]}(\lambda_i)=\sum_{k=0}^{K}\alpha_{[p]}^{(k)} (T_k(L)R_{[p]})^T_m \\
\end{equation}
where $R_{[p]}=[\bm{r}_{[p]}(v_1),\bm{r}_{[p]}(v_2), \dots, \bm{r}_{[p]}(v_n)]^T$. 

Or equivalently,
\begin{equation}
	\bm{h}_{[p]}(v_m)= c_{mm}\bm{r}_{[p]}(v_m) + \sum_{v_o \in N^K(v_m) } c_{mo}\bm{r}_{[p]}(v_o),   \label{eq.eqamp.1}
\end{equation}
with
\begin{equation}
	c_{mo}=\sum_{k=0}^{K}\alpha_{[p]}^{(k)} (T_k(L))_{m,o}  \quad \forall m, o \in \{1,2, \dots, n \}. \label{eq.eqamp.2}
\end{equation}

Eq.~\eqref{eq.eqamp.1} and Eq.~\eqref{eq.eqamp.2} imply that the filtering strategy corresponds to a message passing scheme within a $K$-hop neighborhood.  In other words, the adaptive filter in each subspace equivalently defines a $K$-hop message passing scheme that is learned from the data and exploits the multi-hop topological information of graphs through polynomials of the graph Laplacian. Signal information is also taken into consideration through the learnable parameters $\{\alpha_{[p]}^{(k)}\}$ as well as in the subspace projection step. More importantly, it permits to represent features that do not only have ``low-pass'' properties and learns the specific frequency components in a data-driven manner.

\subsection{Diversity Regularization for Filter Banks}
The filters constituting a filter bank should ideally have diverse frequency responses so that the signal is decomposed through filtering into a series of signals with different frequency characteristics. In GSP, the filters are usually band-pass and divide the spectrum into different parts. Considering that strict band-pass filters are difficult to fit through polynomial functions and that the filter bank is used here to extract useful representation rather than reconstruct input signals, we relax this strict band-pass requirement and rather target filters with diverse frequency responses, which we call ``diversity condition''. With the filter $\hat{g}_{[p]}(\lambda)$ given as a $K$-order polynomial function, the regularization on the filter is imposed on the respective polynomial coefficients $\{ \alpha^{(k)}_{[p]} \}_{p,k}$ in Eq.~\eqref{e.filter}. To achieve the diversity condition, we regularize the polynomial coefficients to be well distributed in the parameter space. Considering that the distances of the coefficient vectors of two scaled filters may still be large in terms of the Euclidean distance, we thereby take the cosine distance to measure the distance between the polynomial coefficients of filters.  Specifically, the regularization term is: 
\begin{equation}\label{e.diverse}
	\Omega(\alpha)=\max_{p \neq q} \frac{|<\bm{\alpha}_{[p]}, \bm{\alpha}_{[q]}>|}{\|\bm{\alpha}_{[p]} \|_2 \|\bm{\alpha}_{[q]}\|_2},
\end{equation}
where $\bm{\alpha}_{[p]}=[\alpha_{[p]}^{(0)}, \alpha_{[p]}^{(1)}, \dots, \alpha_{[p]}^{(K)}]^T$. The max function reflects the maximum similarity between the polynomial coefficients of any pair of filters. Through minimizing Eq.~\eqref{e.diverse}, the most similar filters will have different orientations in the parameter space. Thus, all the pairs of filters tend to be different. When $\{T_k(L)\}_k$ is an orthogonal basis such as Chebyshev polynomials, the diversity of  polynomial coefficients $\{\bm{\alpha}_{[p]}\}_p$ implies that filters defined as Eq.~\eqref{e.filter} are different in the frequency domain. More intuitively, the message passing schemes in the spatial domain induced by the filters are different with diverse  $\{\bm{\alpha}_{[p]}\}_p$, as presented in Eq.~\eqref{eq.eqamp.1} and Eq.~\eqref{eq.eqamp.2}. 

We can note that, if the filter is defined on the basis composed of rectangular pulse functions (not the case in this paper), \emph{i.e.,}
\begin{equation}
	T_k(\lambda)=
\begin{cases}
1& \frac{2k}{K+1} \leq \lambda < \frac{2(k+1)}{K+1}\\
0& \text{others}
\end{cases}.
\end{equation}
the ideal subband filter banks, whose filters have different passbands in GSP, just corresponds to the optimal solution to the regularization with $\Omega(\alpha)=0$ in  Eq.~\eqref{e.diverse}, when $K \geq s$.

With $T_{\Theta}(\mathcal{G}, Y)$ generally representing a target function, the objective function is then formulated as 
\begin{equation}\label{e.obj}
	\min_{\Theta}\ T_{\Theta}(\mathcal{G}, Y) + \gamma \ \Omega(\alpha),
\end{equation}
where $Y$ indicates ground truth labels, $\Theta$ denotes the parameter set including $\{\bm{\alpha}_{[p]}, W_{[p]}, \bm{b}_{[p]}\}_{p=1}^s$,  and $\gamma$ is a hyperparameter to adjust the contribution of regularization term. Like most popular graph convolution operators, BankGCN can be optimized via gradient-based methods together with modules such as graph pooling operators in GCNs. It achieves linear computational complexity with $O(K|\mathcal{E}|d)$ and constant learning complexity, similarly to most existing MPGCNs.

\section{Experiments}\label{sec.eval}
In this section, we evaluate the proposed BankGCN in graph classification tasks on several benchmark datasets, and compare it with several popular GCNs. 

\subsection{Experimental Settings}

\begin{table}[tp]
\centering
 \caption{Dataset statistics and properties (L indicates node categorical features and A denotes node attributes).}\label{t:0}
 \resizebox{.99\textwidth}{!}{  
\begin{tabular}{l|ccccccccc}
	\toprule
	Method &ENZ & D\&D &  PROT & NCI1   &NCI109&MUTA& FRAN & CIFAR-10&Ogbg-molhiv\\
	\midrule
		Avg $|\mathcal{V}|$ &32.63&284.32&39.06&29.87&29.68&30.32&16.90&117.63&25.51\\
		Avg $|\mathcal{E}|$ &62.14&715.66&72.82&32.30&32.13&30.77&17.88&564.86&27.47\\
		Node feature&L+A&L&L&L&L&L&A&A&A\\
		Dim(feat) &3+18 &89 &3 &37 &38 &14 &780&5&9\\
		\#Classes&6&2&2&2&2&2&2&10&2\\
		\#Graphs &600&1,178&1,113&4,110&4,127&4,337&4,337&60,000 &41,127\\
	\bottomrule
\end{tabular}}
\end{table}

 \textbf{Datasets and data splits.} For TU datasets \citep{KKMMN2016}, we conduct experiments on seven widely used public benchmark graph classification datasets, including ENZYMES, D\&D, PROTEINS, NCI1, NCI109, MUTAGENICITY, and FRANKENSTEIN \footnote{Datasets could be downloaded from https://ls11-www.cs.tu-dortmund.de/staff/morris/graphkerneldatasets}.  We adopt node categorical features (one-hot encoding) and node attributes as node signals,  depending on availability on the datasets. Specifically,  node attributes  are adopted on FRANKENSTEIN,  node categorical features and node attributes (normalized in range $[0, 1]$) on ENZYMES, and  node categorical features on the other datasets.  The statistics and properties of the datasets are  summarized in Table~\ref{t:0}.  According to \citep{DBLP:conf/icml/LeeLK19},  we use stratified sampling to randomly split each dataset into training, validation and test sets with a ratio of 8:1:1. The trained model with the best validation performance is selected for test. In order to alleviate the impact of data partition and network initialization, we conduct 20 random runs with different data splits and network initializations on each dataset, and report the mean accuracy with standard deviation of these 20 test results.

We further adopt two large benchmark datasets, CIFAR-10 \citep{dwivedi2020benchmarking} and Ogbg-molhiv \citep{hu2020open}, in the experiments. The graph version of CIFAR-10 is composed from the superpixels of images, and  RGB intensities and normalized coordinates form node signals. Ogbg-molhiv is a molecule graph dataset, with 9-dimensional node features including atomic number, chirality, and additional atom features. We  divide data on these two datasets in accordance with \citep{dwivedi2020benchmarking} and \citep{hu2020open}, respectively. Similarly, results are achieved with 3 runs on  CIFAR-10 and 10 runs on Ogbg-molhiv in order to alleviate the impact of network initialization. 

\textbf{Network architectures.} In the experiments, we adopt a similar architecture to \citep{xu2018representation}, and the network consists of four convolution layers, one graph-level readout module and one prediction module for all the datasets. Specifically, the graph convolution layer is designed as introduced in Section~\ref{sec.method} or defined by various baseline models. An $l_{2}$ normalization function is utilized in each convolution layer to stabilize and accelerate the training process, as in \citep{hamilton2017inductive}. A graph-level readout module then aggregates the graph features from all the convolution layers  to generate the graph representation $h_{\mathcal{G}}$. It consists of node-wise mean and maximum operators (or mean operator on Ogbg-molhiv as in \citep{hu2020open}), represented by $\omega(\cdot)$.
\begin{equation}
    h_{\mathcal{G}}={\rm Concat}(\omega( X_{l})| l=1,2,3,4).
\end{equation}
Finally, a prediction module composed of a linear fully connected layer and a softmax layer makes the prediction of the category of the input graph. The cross-entropy (represented as $T_{\Theta}(\mathcal{G}, Y)$) is adopted as the loss function, which together with the regularization $\Omega(\alpha)$ for diversity condition composes the objective function:
\begin{equation}
	T_{\Theta}(\mathcal{G}, Y)  + \gamma \ \Omega(\alpha)
\end{equation}
where $\Theta$ denotes all the free parameters and $Y$ indicates ground truth categorical labels.

In more details,  we consider two versions of the architectures in the experiments in order to evaluate the models under the same number of features and the same number of parameters, respectively. For the first case,  all of models have the same number of feature maps per hidden layer, whereas models are composed of nearly the same number of parameters per hidden layer for the latter case. We consider the first version with TU datasets in Section~\ref{exp.tu} with 64 feature maps per hidden layer, and the second version in Section~\ref{exp.real} with nearly 16,500 and 65,800 free parameters per layer for all the models on CIFAR-10 and Ogbg-molhiv, respectively.

 \textbf{Configurations.} We implement the proposed models in Pytorch \citep{paszke2017automatic} with geometric package \citep{Fey/Lenssen/2019}, and optimize all of the models with the Adam optimizer \citep{DBLP:journals/corr/KingmaB14} on workstations with GPU GeForce GTX 1080 Ti for TU datasets as well as CIFAR-10 and with GPU RTX 2080 Ti  for Ogbg-molhiv.  In more details, for TU datasets, the learning rate is 0.001 and the batch size is 64. The number of training epochs is set as 500, and early stopping is employed with patience 30.  Finally, we obtain the following optimal  hyper-parameters through grid search: weight decay $\in \{0, 1e^{-5}, 1e^{-4}\}$ and  $\gamma \in \{0,  0.1, 10\}$.
 
 For CIFAR-10 and Ogbg-molhiv,  the batch size is increased to 256 due to their large scale.   Similarly to \citep{dwivedi2020benchmarking}, we adopt a dynamic learning rate that is initialized as 0.001 and decays by 0.1 when validation loss is not improved for 20 epochs until the minimum learning rate $1e^{-5}$.  The number of training epochs is 500 with early stoping (patience 50). The other settings are the same as TU datasets.

For the polynomial basis in the filter design of BankGCN, we adopt the widely used Chebyshev polynomials. 
\begin{equation}
	T_0(\lambda)=1, \  T_1(\lambda)=\lambda, \ T_k(\lambda)=2\lambda T_{k-1}-T_{k-2}.
\end{equation}
The graph Laplacian is adopted as $\tilde{L}=L-I$ for numerical stability,  similarly to \citep{defferrard2016convolutional}.

\textbf{Baselines.} We compare BankGCNs with several state-of-the-art graph convolution methods. For the MPGCNs, we consider GCN \citep{kipf2016semi}, GraphSage \citep{hamilton2017inductive} with mean aggregation,  GAT ($8$-heads) \citep{velikovi2018graph}, and GIN   using SUM-MLP (2 layers) that achieves the best performance in \citep{xu2018how}. With the filters in our method defined as Chebyshev polynomials, it is necessary to compare to its counterpart in spectral graph convolution operators, ChebNets \citep{defferrard2016convolutional}. For fair comparison,  the results of baseline models are obtained with the same configurations as BankGCNs using the public versions provided in the pytorch-geometric package \citep{Fey/Lenssen/2019}. 

\begin{table*}[tp]
\caption{Results on graph classification with 20 runs for different datasets.}\label{t:1}
\vskip 0.15in
\centering
\resizebox{\textwidth}{!}{  
\begin{tabular}{l|ccccccc|c}
\toprule
       &ENZY&DD&NCI1&PROT&NCI109&MUTA&FRAN&\#Para/layer\\
\midrule
GCN&62.75 $\pm$ 5.83&77.75 $\pm$ 3.55&79.00 $\pm$  1.93&74.87 $\pm$  4.08&78.90 $\pm$  1.52&81.34 $\pm$ 1.61&62.21 $\pm$ 2.41&4160\\
GraphSage&66.75 $\pm$ 6.31&75.21  $\pm$ 2.72&80.97 $\pm$  1.87&75.13 $\pm$  4.04&79.54 $\pm$ 2.24&82.30 $\pm$ 1.48&63.91 $\pm$ 1.96&8256 \\
GIN&61.08 $\pm$ 4.92&75.42  $\pm$ 3.31&81.19 $\pm$  2.27&74.91 $\pm$  3.88&80.71 $\pm$ 2.38&81.66 $\pm$ 2.48&68.11 $\pm$ 2.09&8320 \\
GAT &62.67 $\pm$ 7.52&77.50  $\pm$ 2.14&79.43 $\pm$  2.38&75.09 $\pm$  4.05&79.16 $\pm$ 1.85&81.28 $\pm$ 2.20&63.89 $\pm$ 1.53&4288\\
ChebNets ($K=2$)&66.75 $\pm$ 4.79&77.67  $\pm$ 2.91&81.80 $\pm$  2.35&74.64 $\pm$  4.75&81.27 $\pm$ 1.89&82.50 $\pm$ 1.58&68.35 $\pm$ 2.65 &12353\\
\midrule
BankGCN \small ($K=2$, $s=8$)&\textbf{68.00 $\pm$ 5.23}&\textbf{78.14  $\pm$ 2.81}&\textbf{82.06 $\pm$  1.75}&75.67 $\pm$  4.19&81.54 $\pm$ 2.13&\textbf{82.89 $\pm$ 1.61}&67.82 $\pm$ 2.30&4163\\
BankGCN \small ($K=2$, $s=16$)&66.83 $\pm$ 5.19&77.42 $\pm$ 3.50&81.93 $\pm$ 2.15&\textbf{76.12 $\pm$ 5.08}&\textbf{81.62 $\pm$ 1.87}&82.57 $\pm$ 1.61&\textbf{68.43 $\pm$ 1.98}&4163\\
\bottomrule
 \end{tabular}}
 \vskip -0.1in
\end{table*}

\begin{table*}
\caption{Study on the order $K$ of filters and the number of subspaces $s$ per layer. }\label{t:2}
\vskip 0.15in
\centering
\resizebox{\textwidth}{!}{  
\begin{tabular}{l|l|ccccccc|c}
\toprule
     \multicolumn{2}{c}{}   &ENZY&DD&NCI1&PROT&NCI109&MUTA&FRAN&\#Para/layer\\
\midrule
$K=1$& \multirow{4}{*}{$s=8$}&67.17 $\pm$ 5.68&76.99 $\pm$ 2.99&81.02  $\pm$  1.88&\textbf{75.89  $\pm$  5.07}&80.92  $\pm$ 1.66&82.40  $\pm$  1.89&66.95  $\pm$  1.91&4162\\
$K=2$&&\textbf{68.00 $\pm$ 5.23}&\textbf{78.14  $\pm$ 2.81}&82.06 $\pm$  1.75&75.67 $\pm$  4.19&\textbf{81.54 $\pm$ 2.13}&\textbf{82.89 $\pm$ 1.61}&67.82 $\pm$ 2.30&4163\\
$K=3$&&65.75 $\pm$ 5.54&77.75 $\pm$ 2.66&81.85 $\pm$ 1.92&74.96 $\pm$  5.77&80.82 $\pm$ 1.84&82.53 $\pm$ 1.56&68.35 $\pm$ 2.13&4164\\
$K=4$&&65.17 $\pm$ 6.62&77.75 $\pm$ 3.01&\textbf{82.46  $\pm$  1.98}&75.09  $\pm$  4.94&81.17  $\pm$ 2.11&82.26  $\pm$  1.71&
\textbf{68.35  $\pm$ 1.92}&4165 \\
\midrule
\multirow{4}{*}{$K=2$}&$s=1$&63.58 $\pm$ 6.31&76.40 $\pm$ 2.34&80.46 $\pm$ 2.34&74.38 $\pm$ 4.80&79.23 $\pm$ 2.29&82.09 $\pm$  1.51&65.52 $\pm$ 2.44&4163\\
&$s=4$&66.75 $\pm$ 5.61&78.14 $\pm$ 2.81&81.62 $\pm$ 1.84&75.67 $\pm$ 4.61&81.19 $\pm$ 2.08&82.70 $\pm$ 1.63&67.48 $\pm$ 2.09&4163\\
& $s=8$&\textbf{68.00 $\pm$ 5.23}&\textbf{78.14  $\pm$ 2.81}&\textbf{82.06 $\pm$  1.75}&75.67 $\pm$  4.19&81.54 $\pm$ 2.13&\textbf{82.89 $\pm$ 1.61}&67.82 $\pm$ 2.30&4163\\
&$s=16$&66.83 $\pm$ 5.19&77.42 $\pm$ 3.50&81.93 $\pm$ 2.15&\textbf{76.12 $\pm$ 5.08}&\textbf{81.62 $\pm$ 1.87}&82.57 $\pm$ 1.61&\textbf{68.43 $\pm$ 1.98}&4163\\
\bottomrule
 \end{tabular}}
\vskip -0.1in
\end{table*}

\begin{figure*}[tp]
\subfigure[\tiny $K=2, \gamma=0$.]{ 
\centering
  \includegraphics[width=0.18\linewidth]{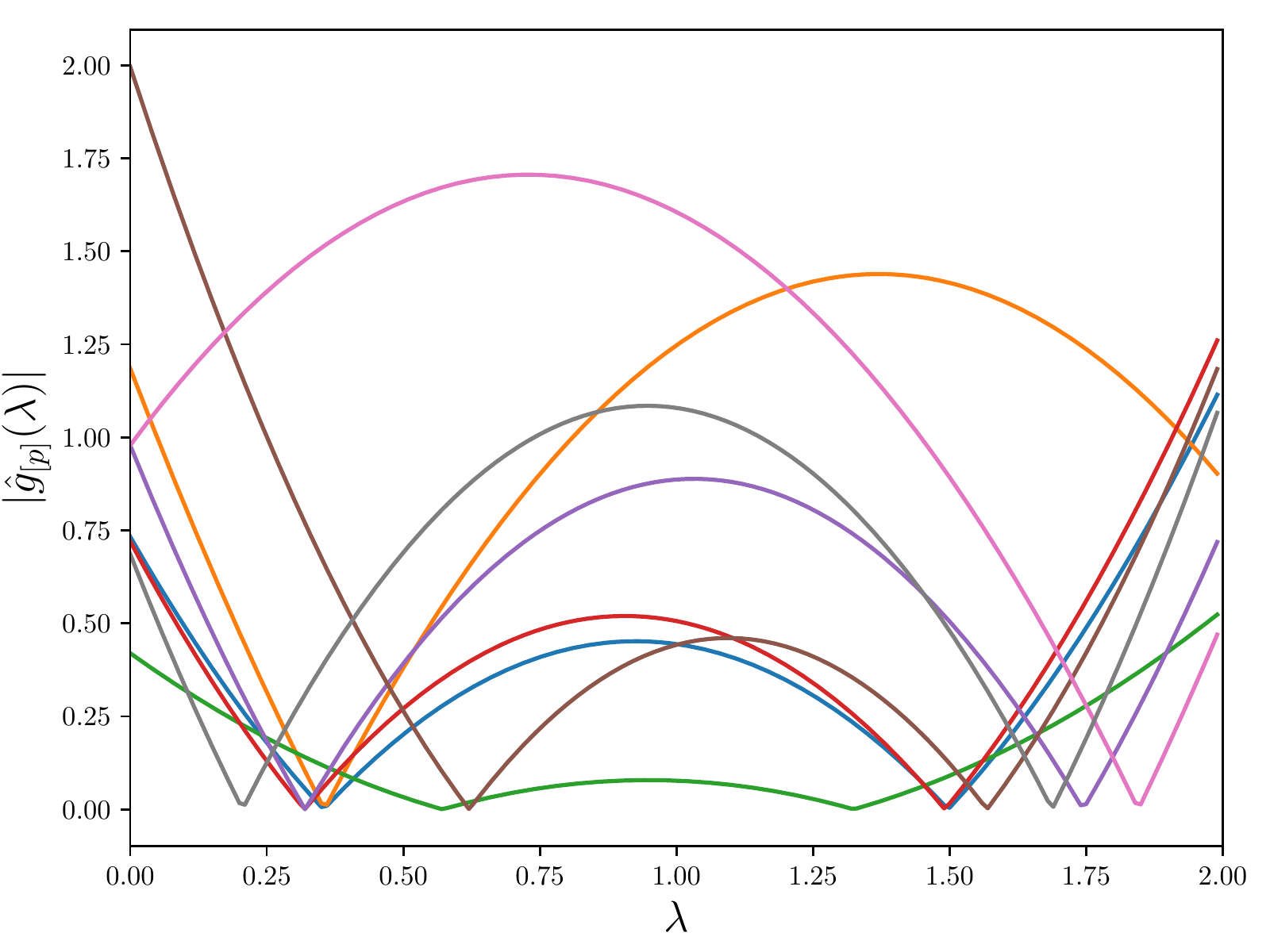}
}
\subfigure[\tiny  $K=2, \gamma=0.1$.]{ 
\centering
  \includegraphics[width=0.18\linewidth]{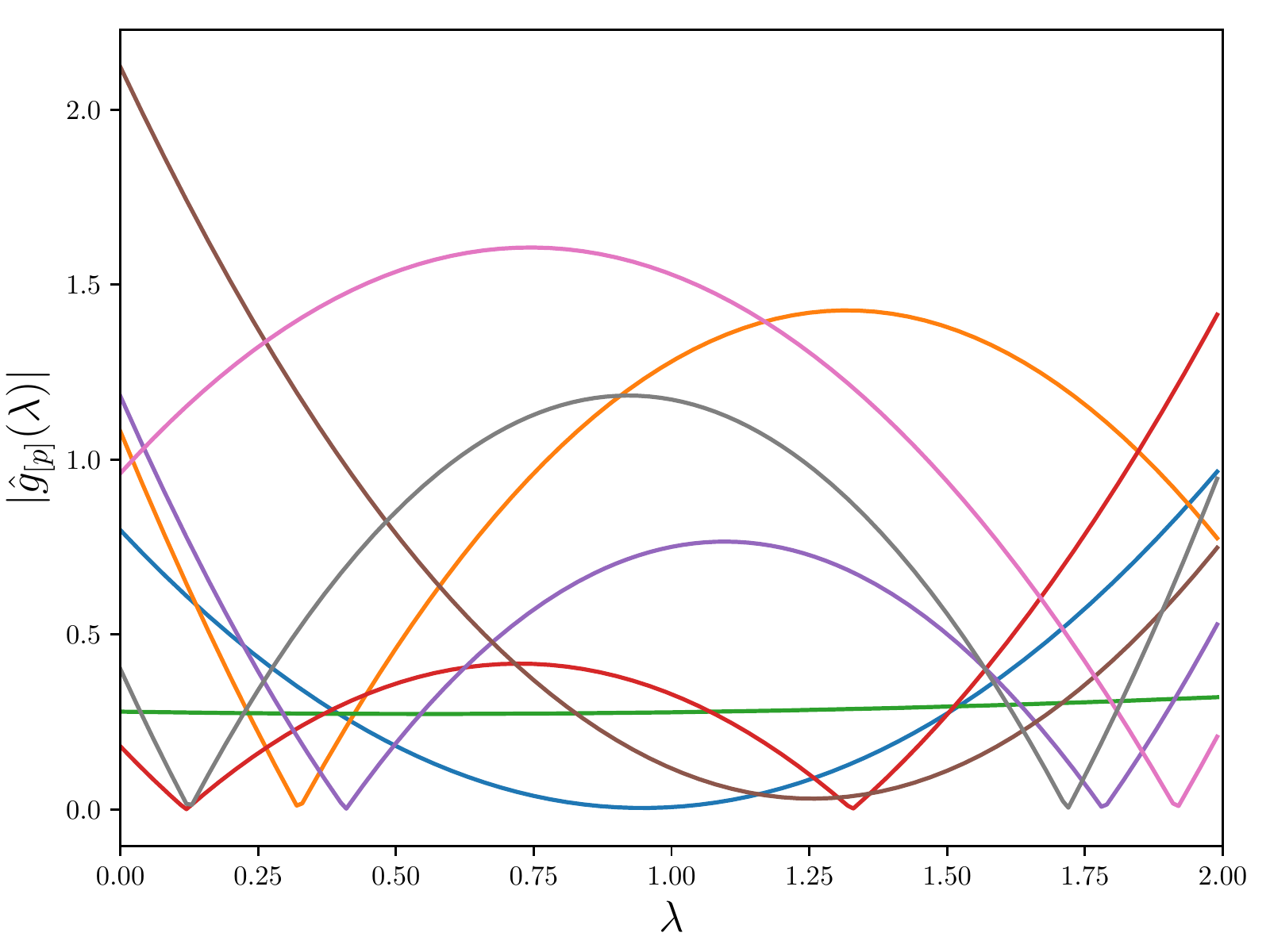}
}
\subfigure[\tiny  $K=2, \gamma=10$.]{ 
\centering
  \includegraphics[width=0.18\linewidth]{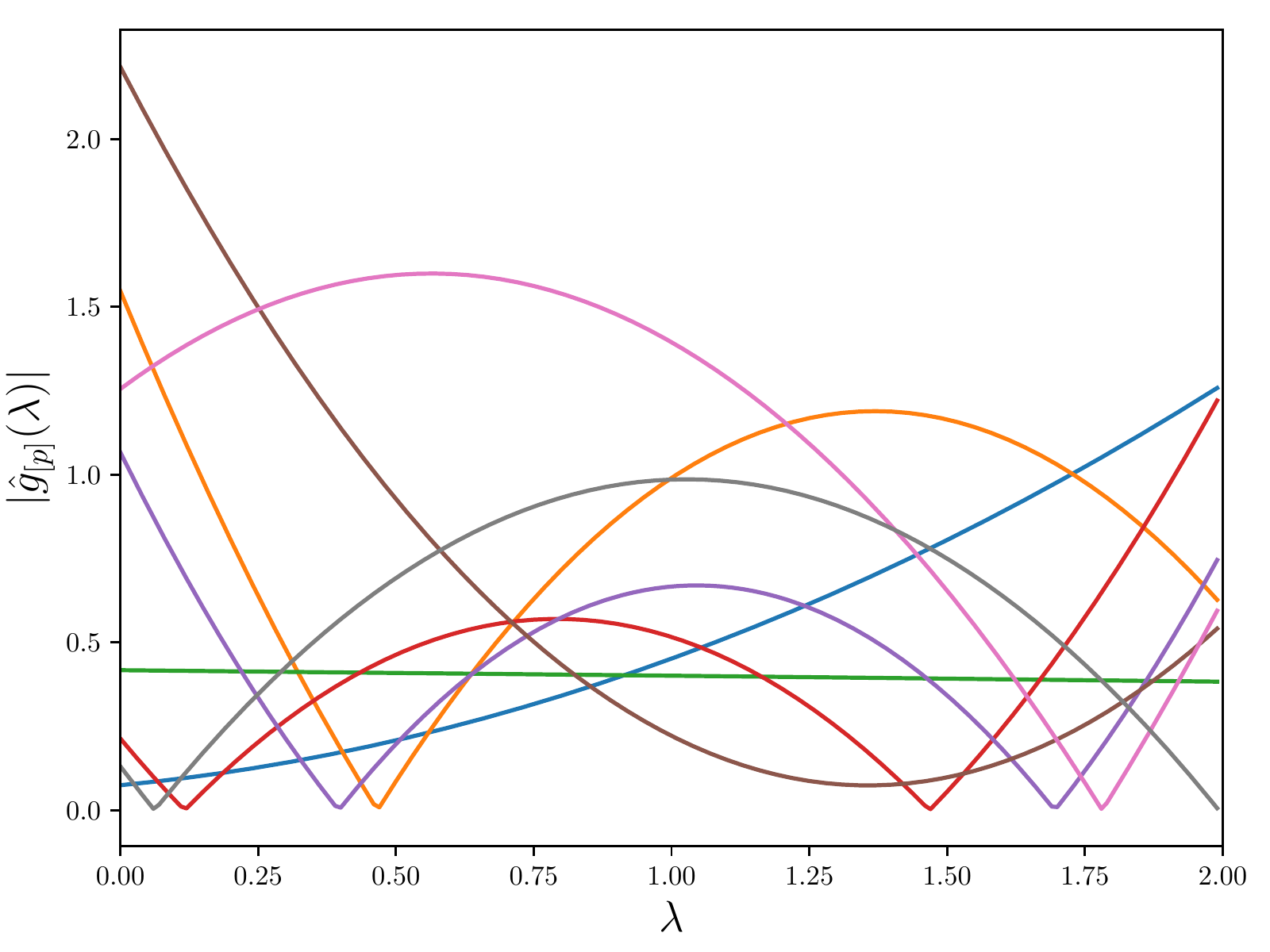}
}
\subfigure[\tiny  $K=3, \gamma=10$.]{ 
\centering
  \includegraphics[width=0.18\linewidth]{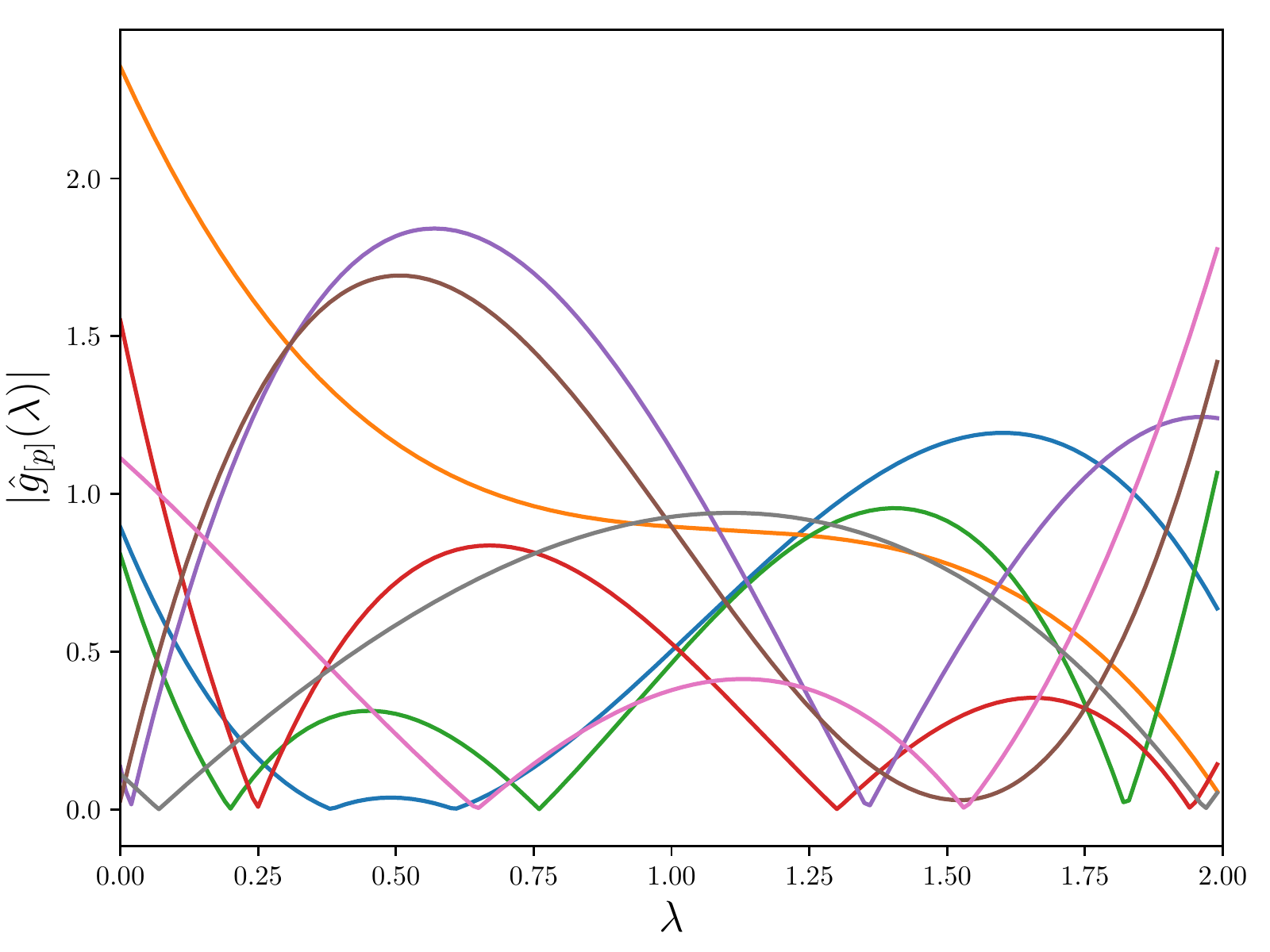}
}
\subfigure[\tiny  $K=2, \gamma=0$.]{ 
\centering
  \includegraphics[width=0.18\linewidth]{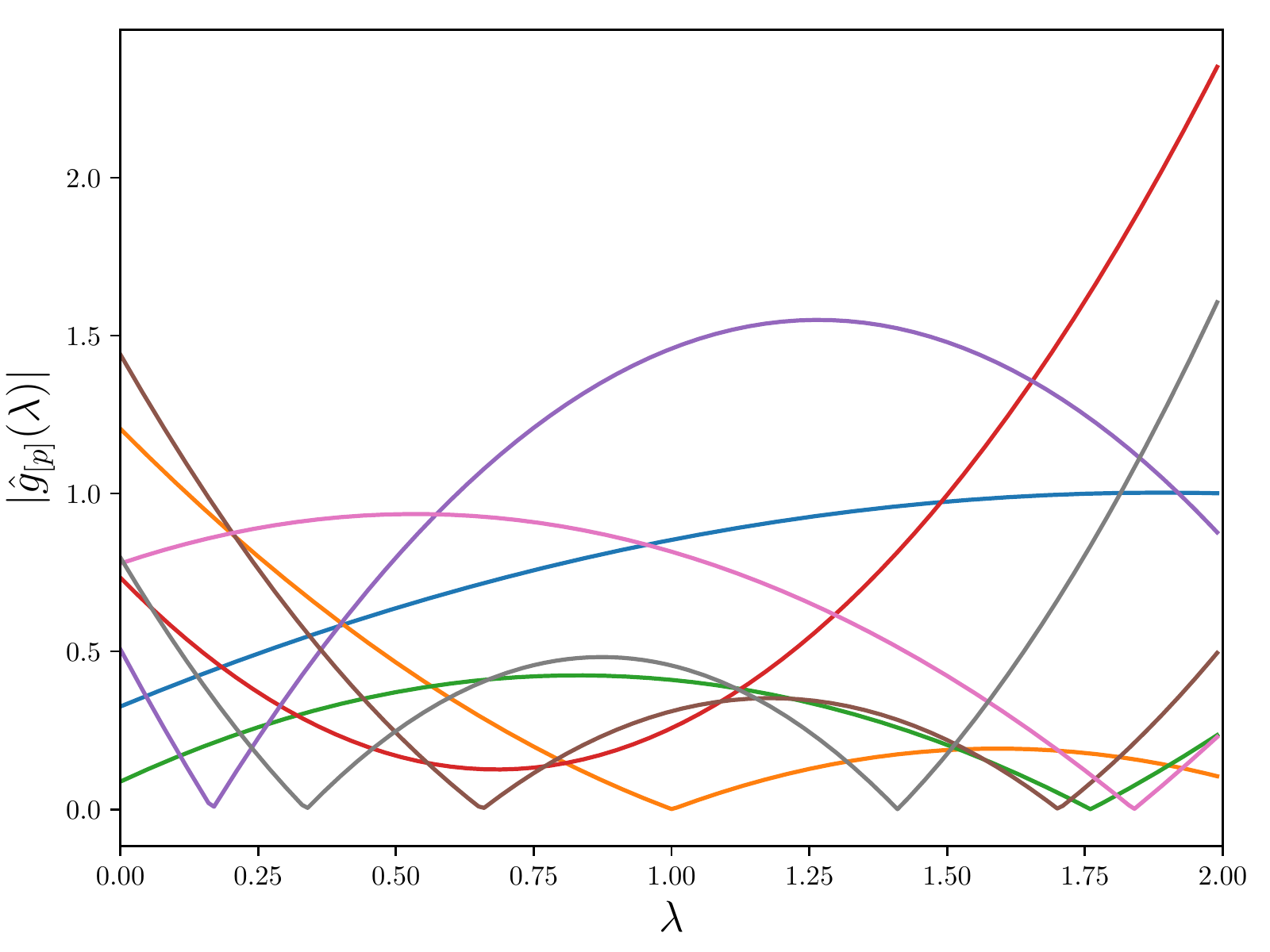}
}
 \caption{Comparisons of the frequency responses of the learned filters of BankGCN ($s=8$) in the first layer of networks. (a) $\sim$ (d) are on NCI109 and (e) on FRANKENSTEIN.} \label{fig.3}
\vskip -0.1in
\end{figure*}

\subsection{Results and Analysis on TU-benchmarks}\label{exp.tu}
As presented in Table~\ref{t:1}, BankGCNs outperforms all the MP baselines with nearly the same number of parameters (BankGCNs has even less parameters than GraphSage and GIN). It also achieves better performance than its spectral counterpart ChebNets with much less free parameters, \emph{i.e.}, about $1/(K+1)$. Fig.~\ref{fig.3} shows that the learned filters do not only focus on low frequency components. Furthermore, filters in a filter bank have different frequency responses. Some suppress high-frequency components, and some focus on middle-frequency frequencies, as demonstrated in Fig.~\ref{fig.3}. With such a bank of filters, BankGCN handles the multi-channel signals flexibly and thereby achieves the superior performance.
 
We then go one step further to evaluate the adaptive filtering capabilities related to the problem Q1 presented in Section~\ref{sec.pb}. As presented in Fig.~\ref{fig.3}, the learned filters have different frequency responses on various datasets as they are adapted to the data characteristics.  Tables~\ref{t:1} and~\ref{t:2} show that the BankGCN ($s=1$) employing one single adaptive filter still outperforms GCN with `low-pass' filtering on most datasets. These validate the benefits of adopting adaptive filters to flexibly capture the frequency characteristics of data.

\textbf{Study on the order of filters.} The order of polynomials determines the function space of filters.  In the frequency domain, as demonstrated in Fig.~\ref{fig.3}, the bandpass property of filters can be better realized with larger $K$ but it will overfit the spectrum of training data. In the graph domain, the value of $K$ corresponds to the neighborhood range to aggregation information, and large $K$ will impact the locality of signals.  Thereby, a tradeoff is needed. As shown in Table~\ref{t:2}, BankGCNs with $K=2$ achieves the best performance on most datasets. We notice that for the cases with complex node signals, like FRANKENSTEIN with node attributes with $780$ channels, relative larger  $K$ is needed in order to exploit their various frequency characteristics; and smaller $K$ is preferred on simple datasets, such as the PROTEINS dataset with $3$-channel node category features.

\begin{table*}[tp]
\caption{Ablation study on the diversity regularization with BankGCN ($K=2$, $s=8$).}\label{t:3}
\vskip 0.15in
\centering
\resizebox{\textwidth}{!}{  
\begin{tabular}{ll|ccccccc}
\toprule
      &   &ENZY&DD&NCI1&PROT&NCI109&MUTA&FRAN\\
\midrule
&$\gamma=0$     &65.83 $\pm$ 6.66&77.03 $\pm$ 4.08&81.89  $\pm$  1.95&75.36  $\pm$  4.68&81.03  $\pm$  1.95&82.44  $\pm$  1.69&\textbf{67.82  $\pm$  2.30}\\
&$\gamma=0.01$&66.50 $\pm$ 6.39&77.54 $\pm$ 3.21&81.63  $\pm$  2.33&75.49  $\pm$  4.07&81.27  $\pm$  2.16&82.87  $\pm$  1.94&67.71  $\pm$  1.96\\
&$\gamma=0.1$  &\textbf{68.00 $\pm$ 5.23}&\underline{78.09 $\pm$ 2.18}&\textbf{82.06  $\pm$  1.75}&75.67  $\pm$  4.19&\underline{81.34  $\pm$  1.92}&82.83  $\pm$  1.87&67.68  $\pm$ 2.10 \\
&$\gamma=1$     &66.42 $\pm$ 6.20&77.97 $\pm$ 3.57&81.81  $\pm$  1.97&\textbf{76.29  $\pm$  4.84}&81.00  $\pm$  2.21&\textbf{82.95  $\pm$  1.44}&\underline{67.76  $\pm$  1.65}\\
&$\gamma=10$   &66.75 $\pm$ 5.90&\textbf{78.14 $\pm$ 2.81}&\underline{82.06  $\pm$  2.10}&75.31  $\pm$  4.64&\textbf{81.54  $\pm$  2.13}&\underline{82.89  $\pm$  1.61}&67.20  $\pm$ 1.67 \\
&$\gamma=100$ &\underline{67.08 $\pm$ 5.42}&77.75  $\pm$ 3.26&81.81  $\pm$ 2.18&\underline{75.98  $\pm$  4.40}&81.14  $\pm$  1.94&82.44  $\pm$  1.71&67.52  $\pm$ 1.83 \\
\bottomrule
 \end{tabular}}
 \vskip -0.15in
\end{table*}

 \textbf{Study on the number of filters.}  Furthermore, we study the impact of the number of filters in the filter bank (equivalently $s$, the number of subspaces),  on the classification performance to show the benefits of subspace projection described in the question Q2 in Section~\ref{sec.pb}.  With a group of filters, the ability of convolution operators to handle information  is enhanced.  As presented in the bottom part of Table~\ref{t:2}, as $s$ increases from $1$ to $8$, the performance of  BankGCN is improved on most datasets. This validates the necessity of using more than one filter. With $s$ further increased into $16$, the performance is degraded on several datasets. Given that the total dimension of all the subspaces is fixed, the dimension of each space decreases and the representation capacity of each subspace probably declines with the growth of $s$.

\begin{table}[t]
\renewcommand{\baselinestretch}{1.0}
\renewcommand{\arraystretch}{1.0}
\renewcommand{\abovecaptionskip}{0pt}
\centering
 \caption{Classification accuracy on  CIFAR-10 and Ogbg-molhiv (no edge attributes) datasets. }\label{t:4}
\begin{tabular*}{0.99\columnwidth}{@{\extracolsep{\fill}}l|cc|c|c} 

	\toprule
	Method &CIFAR-10 &\multicolumn{2}{c|}{ CIFAR-10 (1000)}& Ogbg-molhiv\\ 
	          & Acc           & Acc & Decrease & ROC-AUC \\
	\midrule
    GCN&55.64 $\pm$ 0.11&36.47 $\pm$ 0.31 &\small-34.5\% &75.18 $\pm$ 1.85\\
    GraphSage&63.51 $\pm$ 0.40&40.03 $\pm$ 0.56 &\small-37.0\%&75.39 $\pm$ 1.64\\
    GIN&50.04 $\pm$ 0.06&31.97 $\pm$ 0.20 &\small -36.1\%&71.52 $\pm$ 1.45\\
    GAT&60.34 $\pm$ 0.19&36.08 $\pm$ 0.04 &\small-40.2\%&75.08 $\pm$ 0.39\\
    ChebNets ($K=2$)&64.33 $\pm$ 0.14&39.46 $\pm$ 0.75 &\small-38.7\%&74.69 $\pm$ 2.08\\
    ChebNets ($K=3$)&63.62 $\pm$ 0.23&37.91 $\pm$ 0.40 &\small-40.4\%&73.17 $\pm$ 1.57\\
    \midrule
   BankGCN($K=2, s=16$)&\textbf{66.17 $\pm$ 0.34}&42.82 $\pm$ 0.33 &\small-35.3\%&\textbf{77.95 $\pm$ 1.96}\\
   BankGCN($K=3, s=16$)&66.00 $\pm$ 0.51&\textbf{42.95 $\pm$ 0.49} &\small-34.9\%&75.72 $\pm$ 1.45 \\
	\bottomrule
\end{tabular*}
\vskip -0.1in
\end{table}

\textbf{Ablation Study.}  We further evaluate the effect of the diversity regularization proposed in Section 4.4. In Table~\ref{t:3}, we consider the values of $\gamma \in \{0, 0.01, 0.1, 1, 10, 100\}$ to adjust its contribution in the objective function  Eq.~\eqref{e.obj}.  The regularization improves the classification performance on almost all the datasets. Mostly, the best performance is achieved with $\gamma=0.1$ or $\gamma=10$, and thereby we consider $\gamma \in \{0, 0.1, 10\}$ in  adjusting the contribution of regularization in the experiments. On the FRANKENSTEIN dataset whose signal is composed of $780$-channel attributes, the regularization is not helpful. We infer that the information in such high-channel signal is complex enough to induce different filters, as presented in Fig.~\ref{fig.3}(e).  Fig.~\ref{fig.3}(a)-(c) show the learned filters in a filter bank with regularization present better diversity in terms of frequency response, than those without regularization. For example, the filters denoted by blue and red in Fig.~\ref{fig.3}(a) are with similar frequency responses, while they are more diverse in Fig.~\ref{fig.3}(b) and Fig.~\ref{fig.3}(c). This is further verified by the maximum similarity scores of the polynomial coefficients that define the filters $\Omega(\alpha)=0.997$, $0.744$, and $0.649$ (computed as Eq.~\eqref{e.diverse}) for Fig.~\ref{fig.3}(a)-(c), respectively.

\subsection{Results and Analysis on CIFAR-10 and Ogbg-molhiv }\label{exp.real}
BankGCNs achieves the best performance on both CIFAR-10 and Ogbg-molhiv, as presented in Table~\ref{t:4}, with all the models having a similar number of free parameters per hidden layer.  Furthermore, we construct a reduced CIFAR-10 (1000) dataset by taking 100 graphs per category to form the training set, while maintaining the validation and testing set. BankGCNs is still superior on the reduced CIFAR-10 and is among the models with the least performance loss compared with the full dataset.  Together with ROC-AUC being a measure of the generalization ability of a model, BankGCN performs well in the sense of generalization, especially when compared with its spectral counterpart ChebNets.

\section{Conclusion}\label{sec.con}
In this paper,  we propose a novel graph convolution operator, termed BankGCN, constructed on an adaptive filter bank to process heterogeneous information of graph data. The filter bank is equivalent to a group of learnable message passing schemes in $K$-hop neighborhoods. Further with subspace projection, BankGCN presents powerful capacity to adaptively handle information of diverse frequency components with significantly less parameters than its competitors, and achieves excellent performance on graph classification tasks. An interesting direction for future research resides in discussing the capacity of the proposed graph convolution operator in terms of graph isomorphism test. It may also be promising to employ BankGCN in a variety of tasks on non-Euclidean data like 3-D point cloud classification and segmentation. 

\bibliography{egbib}
\bibliographystyle{abbrvnat}

\end{document}